\documentclass[conference]{IEEEtran}

 \usepackage[english]{babel}
 
\usepackage{multirow, makecell}
\usepackage{tablefootnote}

\usepackage{cite}
\usepackage{amsmath,amssymb,amsfonts}
\usepackage{algorithmic}
\usepackage{graphicx}
\usepackage{textcomp}
\usepackage{xcolor}

\usepackage[hyperfootnotes=true,hidelinks]{hyperref}
\usepackage[capitalise]{cleveref}
\usepackage{array}
\usepackage[binary-units=true]{siunitx}

\usepackage[symbol]{footmisc}

\usepackage[acronym,nomain]{glossaries}
\glsdisablehyper

\newacronym{adas}{ADAS}{advanced driver-assistance systems}
\newacronym{dl}{DL}{Deep Learning}
\newacronym{dnn}{DNN}{Deep Neural Network}
\newacronym{hwa}{HWA}{Hardware Accelerator}
\newacronym{ip}{IP}{Intellectual Property}
\newacronym{pi}{PI}{Performance Indicator}
\newacronym{ops}{OPS}{Operations Per Second}
\newacronym{soc}{SoC}{System on a Chip}
\newacronym{ssd}{SSD}{Single Shot Detector}
\newacronym{uC}{µC}{Microcontroller}

\def\BibTeX{{\rm B\kern-.05em{\sc i\kern-.025em b}\kern-.08em
    T\kern-.1667em\lower.7ex\hbox{E}\kern-.125emX}}

\DeclareRobustCommand*{\IEEEauthorrefmark}[1]{%
	\raisebox{0pt}[0pt][0pt]{\textsuperscript{\footnotesize #1}}%
}

\begin{document}

\title{Bosch Deep Learning Hardware Benchmark}

\author{
	\IEEEauthorblockN{	Armin Runge\IEEEauthorrefmark{1},
						Thomas Wenzel\IEEEauthorrefmark{2},
						Dimitrios Bariamis\IEEEauthorrefmark{2},\\
				 		Benedikt Sebastian Staffler\IEEEauthorrefmark{3},
				 		Lucas Rego Drumond\IEEEauthorrefmark{2} and
				 		Michael Pfeiffer\IEEEauthorrefmark{3}}

	\IEEEauthorblockA{\IEEEauthorrefmark{1}Department of Advanced Digital Technologies, Bosch Corporate Research, Renningen, Germany}
	\IEEEauthorblockA{\IEEEauthorrefmark{2}Computer Vision Lab, Bosch Corporate Research, Hildesheim, Germany}
	\IEEEauthorblockA{\IEEEauthorrefmark{3}Bosch Center for Artificial Intelligence, Renningen, Germany}
		}

\maketitle

\begin{abstract}
The widespread use of \gls{dl} applications in science and industry has created a large demand for efficient inference systems.
This has resulted in a rapid increase of available \glspl{hwa} making comparison challenging and laborious.
To address this, several \gls{dl} hardware benchmarks have been proposed aiming at a comprehensive comparison for many models, tasks, and hardware platforms.

Here, we present our \gls{dl} hardware benchmark which has been specifically developed for inference on embedded \glspl{hwa} and tasks required for autonomous driving.
In addition to previous benchmarks, we propose a new granularity level to evaluate common submodules of \gls{dl} models, a twofold benchmark procedure that accounts for hardware and model optimizations done by \gls{hwa} manufacturers, and an extended set of performance indicators that can help to identify a mismatch between a \gls{hwa} and the \gls{dl} models used in our benchmark.
\end{abstract}

\section{Introduction}
\label{sec:intro}

\begin{figure}[t]
	\centering
	\includegraphics[width=\linewidth]{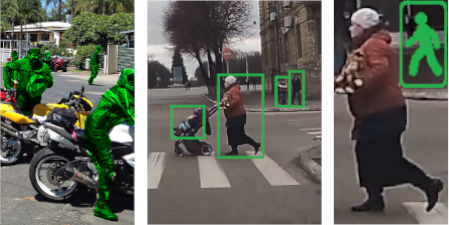}
	\caption{Examples visualizing the three tasks considered in our benchmark (from left to right): semantic segmentation, object detection, and action recognition. In addition to task-specific benchmarks, we include the novel concept of feature extractor benchmarks.}
	\label{fig:task_overview}
\end{figure}

Deep learning hardware benchmarks are of vital importance for the evaluation and comparison of \gls{dl} \glspl{hwa}\footnote{For the sake of brevity, we use the term \gls{dl} \gls{hwa} for any hardware architecture to compute \gls{dl} models. This can also refer to a generic processor like a DSP or Microcontroller ({\si\micro}C).} due to the rapidly evolving number of \gls{dl} models and accelerators. 
We are currently tracking a list of over 200 \glspl{hwa}, which keeps growing on a weekly basis.
A major challenge for customers of these accelerators is the need for a use-case specific evaluation and comparison.
For instance, the requirements of small image classification (e.g.\ ImageNet \cite{ILSVRC15}) vastly differ from those of 4k image semantic segmentation.
This applies to both number of computations and memory accesses as well as to the building blocks used in these \gls{dl} models.
Therefore, Bosch has developed its own \gls{dl} hardware benchmark.
This benchmark focuses on the inference phase of embedded \glspl{hwa} and reflects the particular requirements of computer vision applications, such as those of autonomous driving visualized in \cref{fig:task_overview}.

In recent months, industry standards for \gls{dl} hardware benchmarks appear to become increasingly established.
This is needed in order to cope with the increasing number of both \gls{dl} models and \glspl{hwa} and to keep the effort for both hardware vendors and customers at an acceptable level.
In addition to enabling an evaluation, we see benchmarks as a tool to communicate industry requirements to hardware vendors.

In this paper we describe the reasoning behind our benchmark design, and propose to use this as a starting point for further contributions to public benchmarks and industry standards.
We are convinced that our benchmark contains several novel aspects of interest to the community.
Besides the general model selection, our main benchmark design contributions are the following:
\begin{itemize}
\item A combination of model-level, also referred to as macro-level, benchmarks in addition to feature-extractor benchmarks, hereinafter also referred to as meso-level benchmarks.
\item A twofold benchmark procedure based on unoptimized benchmarks, allowing a direct evaluation of certain \gls{dl} structures and building blocks, in addition to optimized benchmarks, allowing an evaluation of the optimization capabilities in both hardware, but especially also in the corresponding software tools.
\item An extensive list of performance indicators, thus enabling partial cross-validation.
\end{itemize}
However, we will not be able to share results of specific \glspl{hwa} due to confidentiality.

\begin{figure*}
	\includegraphics[width=\linewidth]{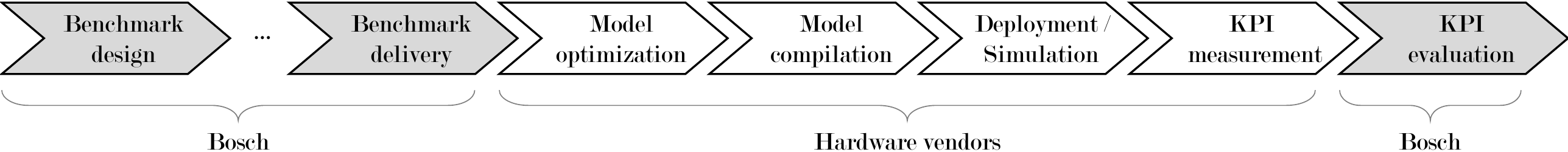}
	\caption{Benchmark roll-out and evaluation procedure that is conducted in a cooperation with each hardware vendor.}
	\label{fig:eval_flow}
\end{figure*}

The remainder of this paper is structured as follows:
we summarize related work in \cref{sec:rel_work} and then give a general overview of design considerations and decisions in \cref{sec:design_consid}.
The benchmark structure and included models are presented in \cref{sec:bench_struc}.
Our evaluation procedure (see \cref{fig:eval_flow}) and the used performance indicators are addressed in \cref{sec:perf_indic} before concluding in \cref{sec:summary}.

\section{Related Work}
\label{sec:rel_work}

As many \gls{dl} hardware benchmarks already exist \cite{Zhang2018}, we provide a brief overview of similar benchmarks in chronological order:

\emph{Fathom} \cite{Adolf2016}, already published in 2016, is one of the first \gls{dl} hardware benchmarks.
It consists of eight models which cover tasks ranging from sentence translation to Atari-playing.
Another pioneer in the field of \gls{dl} hardware benchmarks is Baidu's \emph{DeepBench} \cite{Baidu2016}, which focuses on benchmarking of basic \gls{dl} operations such as convolutions or matrix multiplications.
\emph{DAWNBench} \cite{Coleman2017, Coleman2019}, an end-to-end \gls{dl} training and inference benchmark suite, focuses on image classification and question answering.
The authors recently announced that DAWNBench will stop accepting rolling submissions in favor of MLPerf (see below).
The International Open Benchmark Council, a non-profit research institute, recently published several benchmark suites for different domains, including \emph{Edge AIBench} \cite{Hao2018}, \emph{AIoTBench} \cite{Luo2018}, and \emph{AIBench} \cite{Gao2018}.
The number of benchmarks included in these suites is enormous and the benchmarks cover different granularity levels, different tasks, and domain specific application tests.
Application specific benchmarks for the automotive domain have been developed by EEMBC (\emph{ADASMark} \cite{EEMBC2018}) as well as Basemark (\emph{BATS} \cite{Basemark2019}).
However, as these benchmark suites are focusing on \gls{adas}, the \gls{dl} benchmarks are only a partial aspect of the complete benchmark suite.
Another organization which is rapidly gaining momentum is \emph{MLPerf} \cite{MLPerf}, a collaboration of companies and researchers from educational institutions.
Both a training benchmark \cite{Mattson2019} and an inference benchmark exist \cite{JanapaReddi2019}, which are partly based on the same models.

\section{Motivation and Design}
\label{sec:design_consid}

This section gives an overview of the fundamental design considerations in \gls{dl} hardware benchmark design and motivates the decisions made for the Bosch Deep Learning Hardware Benchmark.

\subsection{Motivation}
A large number of different \gls{dl} hardware benchmarks exist, but each benchmark has unique characteristics and is designed for different domains and evaluation scenarios.
Our focus is on evaluating only the inference performance of \glspl{dnn} on embedded \glspl{hwa}, which considers latency, throughput, accuracy, and other important performance indicators.
We do not consider acceleration of training here, since we assume this is done offline.
Existing \gls{dl} hardware benchmarks were not suitable for our purposes at the time of development, as (1) they did not cover all aspects relevant for us, (2) they were designed for other hardware platforms (e.g.\ GPUs or \si\micro Cs), (3) they were designed for different conditions, e.g.\ focusing on training instead of inference.

The fact that industry wide standards seem to become established is encouraging.
The rapidly increasing number of \gls{dl} models and \gls{dl} hardware architectures leads to a significant increase in the effort required for a fair comparison with state of the art models and datasets for both hardware vendors and customers.
Industry standards offer the chance that these efforts can be bundled.

\subsection{Training vs. Inference}
Most \gls{dl} algorithms involve two consecutive stages: training and inference.
During training, the model is taught to solve a task, e.g.\ $1000$-class image classification, on a training dataset.
During inference, the model predicts the task outcome on unseen data.
Since the model is typically not changed anymore during inference, training and inference can be treated independently and may thus generally be conducted on different hardware with potentially strongly differing characteristics.
In our benchmark we only focus on inference.

\subsection{Granularity}
One fundamental design consideration of every \gls{dl} hardware benchmark is its granularity, i.e.\ at which level benchmarks are defined.
\Cref{tab:granularity} shows an overview of typical benchmark granularity levels.
\emph{Kernel-level} benchmarks, often referred to as micro benchmarks, test single operations, such as convolutions or matrix multiplications.
This enables a direct evaluation of the tested kernel operations and their parameters.
Next are \emph{layer-level} benchmarks, which test individual network layers, providing an analogous evaluation.
The most common benchmark category is at \emph{model-level}, also referred to as macro level.
Such benchmarks are based on complete \gls{dl} models, where the main advantage are measurable model accuracies and effects across layer boundaries, such as layer fusion.

\emph{Sub-model-level} benchmarks are based on several layers, which do not define complete models. 
Specifically, they do not solve any task, which in turn does not allow a task accuracy evaluation such as classification accuracy. 
We refer to them as \emph{meso-level} benchmarks.
In this work we present how to successfully employ meso-level benchmarking to increase the expressiveness of a \gls{dl} hardware benchmark.

Finally, \emph{task-level} benchmarks are not limited to a concrete \gls{dl} model, but only describe a task, e.g.\ object detection in fixed-size images.
Thus, this level incurs the least restrictions, but may require comparing differing \gls{dl} model implementations on different hardware architectures, thus hindering a direct comparison.

\begin{table}[tbp]
\centering
\setlength{\extrarowheight}{2pt}
\begin{tabular}{l l|c|c|c}
& Hardware Arch. & Fixed & Configurable & Programmable\\
\cline{2-5}
& Flexibility & Low & Medium & High\\
\hline
\hline
\multirow{5}{*}{\rotatebox[origin=c]{90}{\parbox[c]{1cm}{\centering Granularity Level}}} & Kernel &  & (x) & x\\
\cline{2-5}
& Layer &  & x & x\\
\cline{2-5}
& Sub-model &  & x & x\\
\cline{2-5}
& Model & (x) & x & x\\
\cline{2-5}
& Task & x & x & x\\
\hline
\end{tabular}
\vspace{1.5mm}
\caption{Typical \gls{dl} hardware benchmark granularities and how they suit to a coarse categorization of \gls{dl} \glspl{hwa}.}
\label{tab:granularity}
\vspace{-2mm}
\end{table}

The above considerations clearly show that the benchmark level has to be chosen carefully depending on the specific goals.
To explain this in more detail, consider three \glspl{hwa}, namely one supporting a fixed \gls{dl} model, a configurable one of medium flexibility and a fully programmable one.
While all \gls{dl} hardware benchmark levels are suitable for the latter, kernel-level benchmarks may be less suitable for the configurable \gls{hwa}, as they might not be supported in an efficient manner or at all.
The fixed \gls{hwa}, which represents the category of ultra low power, in-memory computing based accelerators, is the least flexible and can practically only be tested with task-level benchmarks. 
In this case, prediction accuracy has to be evaluated carefully when comparing \glspl{hwa} that support different models. Model-level benchmarks could also be supported, provided that the tested model is restricted to the native \mbox{(sub-)}model of the \gls{hwa}.

As our benchmark is aiming for accelerators with medium to high flexibility, layer- to task-level benchmarks would be suitable.
However, since we consider optimizations across layer boundaries to be crucial, we decided for a combination of model- and sub-model-level benchmarks (see \cref{sec:bench_struc}).

\subsection{Representation}
Another important aspect of a \gls{dl} hardware benchmark is the representation level of the single benchmarks, i.e. how abstract a benchmark task is defined.
While task-level benchmarks can be described in a textual way, benchmarks of all other levels are usually described in a reference format based on a specific \gls{dl} framework, an exchange format, or an intermediate representation.
Since our primary goal was to enable easy adoption for as many \glspl{hwa} as possible, we decided to use TensorFlow \cite{abadi2016tensorflow}.
However, some hardware vendors converted our models to other frameworks such as Caffe \cite{jia2014caffe}, as their deployment toolchain was optimized for it.
In the future, unified representations such as ONNX \cite{ONNX} could be considered.

\subsection{Modifications and Optimizations}
\label{subsec:benchmark_categories}
Several methods exist to optimize the inference of \gls{dl} models, both regarding hardware (e.g.\ dedicated compression modules), software (e.g.\ kernel pruning), or a combination of both (e.g.\ quantization).
In particular for embedded systems, which usually have strict energy, latency, and throughput requirements, those techniques are highly relevant and almost all dedicated \glspl{hwa} and the corresponding software toolkits support a combination of such techniques.
As several of these modify the \gls{dl} model significantly, they complicate the evaluation of individual hardware features, such as the efficiency of a specific layer structure.
Thus, the question arises, which techniques shall be allowed in a \gls{dl} hardware benchmark.

We classify \gls{dl} hardware benchmarks into four categories:
\begin{enumerate}
\item identical computation, i.e.\ the \gls{dl} model is executed as supplied
\item \label{bm_cat:2} identical computation, but quantization allowed
\item optimization techniques allowed, but without retraining
\item \label{bm_cat:4}optimization techniques allowed including retraining
\end{enumerate}

In our benchmark we allow two of these four categories.
As it is designed to enable an evaluation of certain building blocks of \glspl{dnn}, on the one hand, we are interested in direct results without optimization.
As several embedded \gls{hwa} do not support floating-point operations, we allow only quantization for \emph{non-optimized} results, which corresponds to category \ref{bm_cat:2}.

On the other hand, we are also interested in the aforementioned optimization techniques, and in particular the capabilities of the deployment toolchain.
Since our focus here is on the evaluation of the maximal optimization level, we allow all optimizations techniques, including retraining in the category \emph{optimized}, corresponding to category \ref{bm_cat:4}.
The only restriction is a model-specific tolerated accuracy degradation to avoid unreasonable optimizations.

\subsection{Input Resolution}
Clearly, the resolution of processed images has a major impact on the absolute performance of \glspl{hwa}. Many previous works only considered relatively small images, yet for automotive applications, such as object detection and semantic segmentation, a wide field of view and large sensing range are crucial, which in turn requires a high image resolution. We therefore consider images of Full HD resolution ($1920 \times 1080$px), except for the Action Recognition benchmarks.

\section{Benchmark Structure}
\label{sec:bench_struc}

In this section we describe the general benchmark structure, the ideas behind it, the selected benchmark models, and the used datasets.

\subsection{Benchmark Structure Design}
\label{subsec:bench_struc}
Our proposed benchmark is composed of two distinct, yet strongly related parts. During the last years, deep learning models have become increasingly modular. Many new network architectures have been developed in the course of ILSVRC \cite{ILSVRC15}, such as VGG16 \cite{Simonyan2015}, DenseNet \cite{Huang2017}, up to the recent EfficientNet \cite{Tan2019}. Research has shown that combining parts of the ImageNet-pretrained network with the task-specific structure and loss can yield state-of-the-art results.
We term these two elements of a CNN the \emph{feature extractor} or \emph{backbone} of the model, and the \emph{task-specific head}.
As a consequence many tasks can be approached using a wide variety of well-established feature extractors combined with the task-specific head. 

This is one of the key insights we exploit in our benchmark design. In order to reduce the evaluation complexity, we provide orthogonal feature extractor benchmarks and define the task-specific benchmarks based on a single feature extractor.
This allows e.g.\ comparing results of a MobileNet-based \gls{ssd} \cite{Liu2016} for object detection with a VGG-based \gls{ssd} without the need for conducting this benchmark explicitly.

In the following we describe which particular feature extractors and tasks we selected for our benchmark and motivate our choices. A summary is presented in \cref{tab:bench_struct}.

\subsection{CNN building blocks and feature extractor models}
\label{subsec:building_blocks}
Most modern feature extractors consist of specific building blocks, which are arranged and repeated in a regular pattern \cite{Howard2017} \cite{Szegedy2015}, e.g.\ the \emph{fire-module} \cite{Iandola2016}. We will call them \emph{single-block extractors}. Often each such building block incurs specific requirements on \glspl{hwa}, e.g.\ efficient execution of $1\times1$ convolutions. Many other feature extractors combine several differing building blocks in a beneficial way \cite{Zoph2018} \cite{Zhang2018shufflenet}, which we will call \emph{mixed-block extractors}.

Mixed-block extractors often surpass their single-block predecessors in reported metrics, such as parameter count vs.\ accuracy, or FLOPs vs.\ accuracy. For benchmarking purposes however, mixed-block extractors pose a significant challenge. Execution capability and performance will strongly depend on the least-supported building block, which then produces uninterpretable results for complex feature extractors. Conversely, if single-block extractors are computed efficiently on a \gls{hwa}, mixed-block extractors composed of the involved individual blocks are also highly likely to be computed efficiently. If they do not, the influencing factor is easily identifiable.

In the following, we describe the building blocks and feature extractor models we selected for closer investigation.
Note that there are many more available in the literature.

\subsubsection{Vanilla convolutions}
One of the most commonly used network architecture is the VGG-16 \cite{Simonyan2015}, mainly due to its strong results despite its design simplicity. We therefore select it as a baseline architecture. However, due to its large number of parameters and large image resolution used in the benchmark, we use a smaller variant VGG-16$_{0.25}$ by applying a scaling factor $\alpha=0.25$ to the number of filters of each layer.

\subsubsection{Squeeze and expand}
This block was initially introduced in the Inception \cite{Szegedy2015} architecture.
The SqueezeNet architecture \cite{Iandola2016}, which we selected as the second architecture, then adopted it exclusively and at the time resulted in an excessively low-parameter network. 
It mainly relies on reducing the number of large-filter convolutions in favor of $1 \times 1$ convolutions.

\subsubsection{Inverted residual bottleneck block of depthwise separable convolutions}
With the MobileNet architecture \cite{Howard2017}, depthwise separable convolutions were introduced, replacing standard convolutions by channelwise $2$D convolutions followed by $1 \times 1$ depthwise convolutions, i.e.\ across channels.
Due to the increased memory bandwidth requirement of a naive inference approach, depthwise separable convolutions are more challenging for \glspl{hwa} than squeeze and expand-blocks. 
MobileNet v2 \cite{Sandler2018}, the third feature extractor model of this benchmark, incorporates depthwise separable convolutions in the inverted residual bottleneck block.

\subsubsection{Dense inter-layer connections}
The DenseNet architecture \cite{Huang2017} introduced dense inter-layer connections resulting in strong implications on the required memory management.
everal successors, such as SparseNet \cite{Zhu2018}, which is our fourth feature extractor model, relaxed these requirements by reducing the number of connections significantly while preserving the original network performance.

\begin{table*}[tbp]
\centering
\setlength{\extrarowheight}{2pt}
\begin{tabular}{p{.2\linewidth}| p{.25\linewidth}| p{.1\linewidth}| S[table-format=3.0]| S[table-format=2.1]}
	\textbf{Network Architecture} & \textbf{Building Block} & \textbf{Cutoff Layer} & \textbf{Params} & \textbf{GMAC @ Full HD} \\
	\hline
	\hline
	VGG-16$_{0.25}$ & Vanilla convolutions & conv5\_3 & 921k & 40.3 \\
	\hline
	SqueezeNet & Squeeze \& expand & fire9\_concat & 722k & 11.9 \\
	\hline
	MobileNet\_v2$_{1.0}$ & Inv. bottleneck, dw convolutions & block12\_add & 531k & 8.7 \\
	\hline
	SparseNet-40 & Inter-layer connections & activation\_40 & 723k & 38.5 \\
	\hline
	\hline
	VGG-16$_{0.25}$-FCN 8s & Skip connections, upsampling & - & 924k & 40.6 \\
	\hline
	VGG-16$_{0.25}$-SSD & Multi-head output & - & 923k & 40.4 \\
	\hline
	VGG-16$_{0.25}$-ActRec & LSTM & - & 1378k & 0.2 \tablefootnote{Number of MAC operations is based on single time steps and single patches of size $120 \times 80$px.} \\
	\hline
\end{tabular}
\vspace{1.5mm}
\caption{Benchmark structure overview. Top section summarizes feature extractor benchmarks, bottom section task-specific ones.}
\label{tab:bench_struct}
\vspace{-2mm}
\end{table*}
\subsection{Task-specific models}
\label{subsec:task_specific}
We selected the VGG-16$_{0.25}$ as a common feature extractor for all task-specific benchmarks due to the simplicity of its architecture, which is supported by the vast majority of the \glspl{hwa}. This allows capturing the performance characteristics of the task-specific heads, without the risk of an inefficiency in the feature extractor affecting the results.

The tasks, which are illustrated in \cref{fig:task_overview}, were selected based on their relevance for automotive applications, while also paying attention to the differentiation of their building blocks as presented below:

\subsubsection{Semantic Segmentation}
This task usually involves skip-connections and upsampling as building blocks, in order to provide an output at the same resolution as the input and improve the edge precision. For this benchmark, we opted for the fully convolutional network architecture FCN-8s as presented in \cite{Long2015}, using VGG-16$_{0.25}$ as a feature extractor.
A possible challenge for \glspl{hwa} is the memory bandwidth requirement due to long skip-connections, upsampling, and the high output resolution.

\subsubsection{Object Detection}
Several approaches have been proposed for object detection, some decomposing the problem into generating object proposals and then performing the detection \cite{Ren2015} and other directly generating the detections in one step \cite{Liu2016}.
Due to their favorable efficiency, we opted for the latter in the form of a \gls{ssd}.
This task specific head provides outputs in multiple scales, each consisting of a classification and a regression part.

\subsubsection{Action Recognition}
We rely on the CNN-LSTM architecture proposed in \cite{Donahue2015} for the task of action recognition of pedestrian patches.
Each pedestrian patch is resized to a fixed size, processed by the VGG-16$_{0.25}$ feature extractor and the resulting features are processed by fully connected layers, before they are fed into LSTM units.

\subsection{Datasets}
\label{subsec:datasets}
We selected the following datasets for the task-specific benchmarks, as they are available under permissive licenses: 
\begin{itemize}
\item{Semantic segmentation:} a subset of MS\ COCO \cite{Lin2014} with the two classes \emph{person} and \emph{background}, resized to Full HD resolution.
\item{Object detection:} JAAD dataset \cite{Kotseruba2016} for pedestrian detection.
\item{Action recognition:} JAAD dataset with cropped pedestrian patches, whose action was classified into standing or walking. Patches were resized to $120 \times 80$px.
\end{itemize}
These datasets were used by us as well as the hardware vendors for the applicable steps of the evaluation procedure (see \cref{fig:eval_flow}), namely training the models, optionally optimizing them with retraining in the case of optimized benchmarks (see \cref{subsec:benchmark_categories}, category 4) and evaluating the task accuracies.

\section{Evaluation and Performance Indicators}
\label{sec:perf_indic}

In this section we describe our general evaluation procedure, the evaluated \glspl{pi}, and how we validate the \glspl{pi} received from hardware vendors.

\subsection{Evaluation Procedure}

Evaluating a hardware architecture using this benchmark is a multi-stage process (see \cref{fig:eval_flow}).
It usually begins with the transfer of the benchmark to the hardware vendor.
The hardware vendor executes the benchmark on the target platform in 4 steps:
(1) optimizing the models for its target platform. This usually includes quantization (in case of unoptimized execution) and a couple of further optimization steps including retraining (in case of optimized execution).
(2) model compilation using the vendor's toolchain.
(3) model deployment on the target platform, or, in case no silicon exists, execution on a simulation model.
(4) measurement of the \glspl{pi} described in \cref{subsec:pis}.
Finally, the \glspl{pi} are reported back to us and we evaluate and validate them.

\subsection{Performance Indicators}
\label{subsec:pis}

One key element of this benchmark is the list of \glspl{pi} (see \cref{tab:pis}) we request from each hardware vendor.
Besides accuracy, throughput, and latency we inquire values such as bandwidth requirements and memory footprints for each model and benchmark scenario.
The reason for this is twofold.

\begin{table}[tbp]
\centering
\setlength{\extrarowheight}{2pt}
	\begin{tabular}{p{.45\linewidth}| p{.45\linewidth}}
		\textbf{\gls{pi} Description} & \textbf{Unit} \\
		\hline
		\hline
		\multirow{2}{.95\linewidth}{Accuracy\tablefootnote{For task-specific benchmarks only.}} & [mean IoU $\vert$ Log-Avg. Miss-Rate $\vert$ classification accuracy]\\
		\hline
		\multirow{3}{.95\linewidth}{Performance, throughput, and latency at min/max/optimal power operation points} &
		Performance [TOPs]\\
		& Throughput [img/s]\\
		& Latency [ms]\\
		\hline
		\multirow{3}{.95\linewidth}{Avg. / peak bandwidth \& avg. memory footprint for ext.\ \& local memories} &
		Avg. bandwidth [GB/s]\\
		& Peak bandwidth [GB/s]\\
		& Avg. memory footprint [MB]\\
		\hline
		\multirow{2}{.95\linewidth}{Avg. / peak power consumption (static \& dynamic power)} &
		Avg. power consumption [Watt]\\
		& Peak power consumption [Watt]\\
		\hline
		Avg. compute efficiency &
		Utilized processing units [\%]\\
		\hline
	\end{tabular}
	\vspace{1.5mm}
	\caption{Summary of requested hardware performance indicators.}
	\label{tab:pis}
	\vspace{-2mm}
\end{table}

On the one hand, this allows a more in-depth evaluation.
For instance, the reason for a higher than expected latency or lower throughput can be a mismatch between the hardware architecture and the executed \gls{dl} model.
Another reason could be that the accelerator was memory bound due to an adverse communication-to-computation ratio.
Both reasons lead to an underutilization of the available compute units.
Knowledge about the actual required memory bandwidth in addition to the peak bandwidth of the accelerator allows a better assessment of which of the two cases has actually occurred.
This knowledge in turn allows to draw conclusions about what a more suitable model for this accelerator should have looked like.

On the other hand, a large amount of partially redundant data also allows for cross validation.
A very simple example is the relation between the measured compute efficiency, i.e.\ the  percentage of compute units that actually computed something useful, the measured performance, and the accelerator's peak performance.
If a vendor has measured a performance of $t$ \gls{ops} and a compute efficiency of $50\%$, the accelerator's peak performance should be $2t$ \gls{ops}.

\subsection{Pitfalls}
In this section, we address two typical pitfalls in evaluating \gls{dl} hardware benchmark results.

If a benchmark is used to compare \glspl{hwa} of very different performance classes, the results must be interpreted with caution.
Achieved performance, throughput, and latency can be theoretically compared by weighting the results with the accelerators' peak performance\footnote{For the same workload, it is more difficult to fully utilize a high performance accelerator than a low performance accelerator.}.
However, comparing the memory bandwidth requirements is more difficult.
The required bandwidth for a given workload is mainly determined by the accelerator's internal memory capacity.
However, a direct derivation of a quotient like the compute efficiency is not reasonable.
Hence, a model based prediction of the required memory bandwidth based on the \gls{dl} model and the accelerators internal memory capacity can be used.

Another pitfall is the comparison of completely different power numbers.
This is in particular important if the benchmark is used to compare \gls{ip} cores, chips, \glspl{soc}, or even boards.
For instance, board power cannot only be measured for an \gls{ip}. 
However, for chips and \glspl{soc} the power consumption of external memory access should be included, as they are responsible for a substantial part of the total power consumption.
Hence, it is very important to define how and at which level power consumption should be measured.

\section{Summary}
\label{sec:summary}

We presented the Bosch Deep Learning Hardware Benchmark, which focuses on the inference phase of embedded \glspl{hwa} and reflects the requirements of computer vision tasks that are relevant for automated driving.

Our key contributions are reflected in the benchmark design and model selection. In particular, we define a new granularity level for benchmarks, namely meso-level for feature extractors, a twofold benchmark procedure that distinguishes optimization of vendor hardware from software, and an extensive list of performance indicators that allow to easily identify a mismatch between an accelerator and a model.
To this end, we define a carefully selected set of feature extractors and task-specific models.

In the future we plan to become more involved in establishing \gls{dl} hardware benchmark standards and share our models and insights gained from this benchmark with the community.


\bibliographystyle{IEEEtran}
\bibliography{IEEEabrv,literatur}

\end{document}